\documentclass[10pt,leqno,twoside]{article}

\usepackage{annal-latex}
\usepackage{amssymb}
\usepackage{graphicx}
\usepackage{algorithm}
\usepackage{algpseudocode}

\begin{document}
\setcounter{page}{1}
\newcommand\balline{\small Thanh The Van and Thanh Manh Le}
\newcommand\jobbline{\small RBIR using Interest Regions and Binary Signatures}

\vspace{-4cm} \fofej{43}{14}{89}{103}

\vspace{.4cm}

\title{RBIR USING INTEREST REGIONS AND BINARY SIGNATURES}

\author{{\bf Thanh The Van} (Hue, Vietnam)\\[1ex]
				{\bf Thanh Manh Le} (Hue, Vietnam)}

%
\acmclass{H.2.8, H.3.3}
%
%



\commby{J´anos Demetrovics}
\vspace{-2ex}
\recacc{June 1, 2014}{July 1, 2014}

\vspace{-7ex}
\abstract {In this paper, we introduce an approach to overcome the low accuracy of the Content-Based Image Retrieval (CBIR) (when using the global features). To increase the accuracy, we use Harris-Laplace detector to identify the interest regions of image. Then, we build the Region-Based Image Retrieval (RBIR). For the efficient image storage and retrieval, we encode images into binary signatures. The binary signature of a image is created from its interest regions. Furthermore, this paper also provides an algorithm for image retrieval on S-tree by comparing the images' signatures on a metric similarly to EMD (earth mover's distance). Finally, we evaluate the created models on COREL's images.}
\section{Introduction}
It is difficult to find images in a large set of them. A solution is to label the images [1, 2] but it is costly, time-consuming and unfeasible for many applications. Moreover, the labeling process depends on the semantic description of image. So, the image retrieval system which is based on content is developed to extract visual attributes for the description of image content [3]. Some digital image retrieval systems have built, such as QBIC, ADL, DBLP, Virage, Alta Vista, SIMPLIcity, etc.

In recent years, there are some researches regarding CBIR are published, such as image retrieval based on color histogram [1, 2], image retrieval on the base of binary bit string and S-tree [3], extracting objects on the image based on the change of histogram value [5], the similarity image retrieval based on the comparison of feature regions and their similarity relationship on image [6], color image retrieval based on the detection of local regions by Harris-Laplace method [7], color image retrieval based on bit plane and $L^*a^*b^*$ color space [8], changeable color space to query the content of color images [9], etc.

The paper approaches the semantic description of image contents through binary signatures and stores them on a S-tree. The S-tree data structure describes the relationship between binary signatures. So, it describes the relationship in the contents of images. On the base of the description of the semantic relationship of image contents of the S-tree data structure, the paper finds out the similarity images by the content on COREL images [12].

The paper builds the similarity image retrieval on the base of the local interest regions. First of all, it extracts the interest regions using Harris-Laplace method. Then, the paper creates interest regions for image. Basing on these ones, it creates binary signatures and evaluates the similarity of images. In order to speed up the query, the paper presents S-tree to store binary signatures so as to build a similarity image retrieval algorithm on the S-tree. The paper contributes two main sections that reduce the amount of storage space and speed up the query of image objects on a large image database.

The binary signature of images is used to describe the image's content. When compared to GCHs (Global Color Histograms) and CCVs(Color-Coherence Vectors), the binary signature method in VBA (Variable-Bin Allocation) saves over 75\% and 87.5\% in storage overhead, respectively. [3]

\section{Data structure and similarity measure}
\subsection{Binary signature}
In accordance with [4], the binary signature is formed by hashing the data objects, and it has $k$ bits $1$ and $(m - k)$ bits $0$ in the bit chain $[1..m{\rm{]}}$, where $m$ is the length of the binary signature. The data objects and the object of the query are encoded on the same algorithm. When the bits in the signature data object cover completely the ones in the query signature, the data object is a candidate for the query. There are three cases: (1) the data object matches the query: each bit in the ${s_q}$ is covered with the ones in the signature ${s_i}$ of the data object (i.e., ${s_q} \wedge {s_i} = {s_q}$); (2) the object does not match the query (i.e., ${s_q} \wedge {s_i} \ne {s_q}$ ); (3) the signatures are compared and then given a \textit{false drop} result.

\subsection{EMD distance}
Suppose that $I$ is a set of suppliers, $J$ is a set of consumers, ${c_{ij}}$ is the transportation cost from supplier $i \in I$ to consumer $j \in J$, we need to find out flows ${f_{ij}}$ to minimize the total cost $\sum\limits_{i \in I} {\sum\limits_{j \in J} {{c_{ij}}{f_{ij}}} }$ with the constraints [10, 11]: ${f_{ij}} \ge 0,\sum\limits_{i \in I} {{f_{ij}} \le {y_j}} ,\sum\limits_{j \in J} {{f_{ij}} \le {x_i}} ,i \in I,j \in J$, where ${x_i}$ is the provider's general ability $i \in I$, ${y_j}$ is the total need of the consumer $j \in J$. The feasible condition is $\sum\limits_{j \in J} {{y_j} \le \sum\limits_{i \in I} {{x_i}} }$. The EMD distance [10, 11] equals $EMD(x,y) = \frac{{\sum\nolimits_{i \in I} {\sum\nolimits_{j \in J} {{c_{ij}}{f_{ij}}} } }}{{\sum\nolimits_{i \in I} {\sum\nolimits_{j \in J} {{f_{ij}}} } }} = \frac{{\sum\nolimits_{i \in I} {\sum\nolimits_{j \in J} {{c_{ij}}{f_{ij}}} } }}{{\sum\nolimits_{j \in J} {{y_j}} }}$.

Each image in the database is quantized into a fixed number of $n$ colors ${c_1},{c_2},...,{c_n}$. Each color ${c_j}$ is represented by a bit string of length $m$, i.e., $b_1^jb_2^j...b_m^j$ for $1 \le j \le n$, so each image can be described as a sequence of bits $S = b_1^1b_2^1...b_t^1...b_1^nb_2^n...b_m^n$.

The binary signature of image $I$ is $SI{G_I} = B_I^1B_I^2...B_I^n$, where $B_I^j = b_1^jb_2^j...b_m^j$, $b_i^j \in \{ 0,1\} $.
The weight of $B_I^j$ component equals $w_I^j = w(B_I^j) = \sum\limits_{i = 1}^m {(b_i^j \times \frac{i}{m} \times 100)}$. Therefore, we have a weight vector of image $I$ as ${W_I} = \{ w_I^1,w_I^2,...,w_I^n\}$. Let $J$ be the image which needs to calculate the similarity corresponding to the image $I$. We need to minimize the cost to convert color distribution $\sum\limits_{i = 1}^n {\sum\limits_{j = 1}^n {{d_{ij}}{f_{ij}}} }$, where $F = \left( {{f_{ij}}} \right)$ is a color distribution flow matrix between $c_I^i$ and $c_J^j$. Let $D = \left( {{d_{ij}}} \right)$ be a Euclidean distance matrix in RGB color space between $c_I^i$ and $c_J^j$. 

We set ${{\rm{W}}_{\rm{m}}} = \min (\sum\limits_{i = 1}^n {w_I^i} ,\sum\limits_{j = 1}^n {w_J^j} )$ and ${{\rm{W}}_{\rm{M}}} = \max (\sum\limits_{i = 1}^n {w_I^i} ,\sum\limits_{j = 1}^n {w_J^j} )$. The distance $EMD(I,J)$ is the color distribution flows from an image with the color weight as ${{\rm{W}}_{\rm{M}}}$ to an image with color weight as ${{\rm{W}}_{\rm{m}}}$. We have the similarity measure between two images $I$ and $J$ with formula as follows:  $EMD(I,J) = \mathop {\min }\limits_{F = \left( {{f_{ij}}} \right)} \frac{{(\sum\limits_{i = 1}^n {\sum\limits_{j = 1}^n {{d_{ij}}{f_{ij}}} )} }}{{\sum\limits_{i = 1}^n {\sum\limits_{j = 1}^n {{f_{ij}}} } }}$, with $\sum\limits_{i = 1}^n {\sum\limits_{j = 1}^n {{f_{ij}}} }  = {{\rm{W}}_m} = \min (\sum\limits_{i = 1}^n {w_I^i} ,\sum\limits_{j = 1}^n {w_J^j} )$

\subsection{S-tree}
S-tree [3, 4] is a tree with many balanced branches. Each node of the S-tree contains a number of pairs $\langle sig,next\rangle $, where $sig$ is a binary signature and $next$ is a pointer to a child node. The root of S-tree contains at least two pairs and at most $M$ pairs $\langle sig,next\rangle $. All internal nodes in the S-tree can accommodate at least $m$ and at most $M$ pairs $\langle sig,next\rangle $, $1 \le m \le {M \mathord{\left/
 {\vphantom {M 2}} \right. \kern-\nulldelimiterspace} 2}$. The leaves of the S-tree contain image's binary signatures $sig$, along with unique identification $oid$ for those images. The S-tree most height for $n$ signatures is $h = {\log _m}n - 1$.

Building the S-tree is done by the \textit{inserting} and \textit{splitting}. The S-tree only contains a null leaf at the beginning. Then, we insert  signatures into the S-tree. When the node $v$ is full, we split into two nodes. At the same time, the parent node ${v_{parent}}$ is created (if not exist). Simultaneously, two new signatures are inserted into node ${v_{parent}}$.
\begin{figure}[ht]
\centering
\includegraphics[height=2.5cm, width = 7cm]{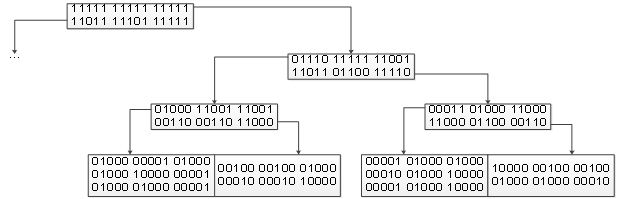}
\includegraphics[height=2cm, width = 12cm]{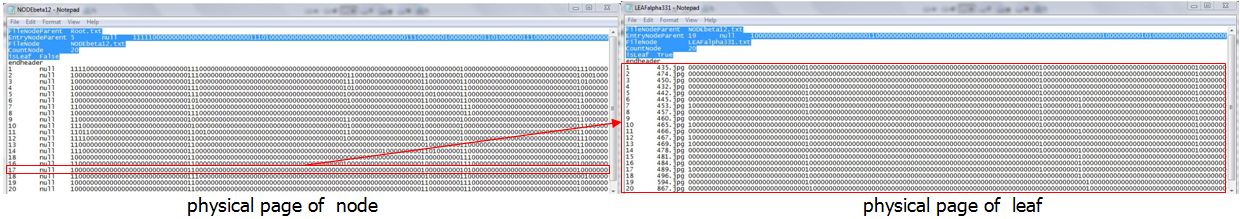}
\caption{A sample of S-tree}
\end{figure}
\begin{figure}[ht]
\centering
\includegraphics[height=3.5cm, width = 8cm]{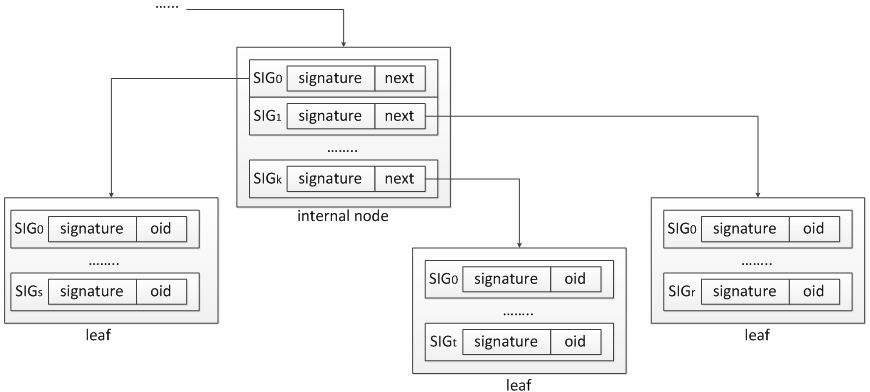}
\caption{A sample of nodes and leaves in S-tree}
\end{figure}

The algorithm creates the S-tree that stores the image's binary signature based on the EMD distance as follows: 
\\\textbf{Algorithm1.} Create the S-tree from the set of signatures
\\\textbf{\textit{Input:}} $S = \{ \langle si{g_i},oi{d_i}\rangle |i = 1,...,n\}$
\\\textbf{\textit{Output:}} The $Stree$.
\begin{algorithmic}[1]
	\State \textit{Step 1.}
			\State $v$ = $Root$;
			\If{$S = \emptyset $}	STOP;
			\Else
				\State Choosing $\langle sig,oid\rangle  \in S;$ $S = S\backslash \langle sig,oid\rangle ;$
				\State Go to Step 2.
			\EndIf
	\State \textit{Step 2.}
		\If{$v$ is $leaf$}
			\State $v = v \cup \langle sig,oid\rangle ;$
			\State UnionSignature($v$);
			\If{$v.count > M$} 
				\State SplitNode($v$);
				\State To go back Step 1;
			\EndIf
		\Else
			\State $EMD(SI{G_0} \to sig,sig) = \min \{ EMD(SI{G_i} \to sig,sig)|SI{G_i} \in v\} ;$
			\State $v = SI{G_0} \to next;$
			\State To go back Step 2;
		\EndIf
	\\\Return $Stree$
\end{algorithmic}
\textbf{Procedure1.} Union of binary signatures at node $v$
\\\textbf{\textit{Input:}} Node $v$
\\\textbf{\textit{Output:}} The $S-tree$ after union of signatures
\begin{algorithmic}[1]
\Procedure{UnionSignature}{${v_{parent}}$}
	\State $s = \bigcup {sig_i^{}} $, with $sig_i^{} \in v;$
	\If{${v_{parent}}! = null$}
		\State $SI{G_v} = \{ SI{G_i}|SI{G_i} \to next = v,SI{G_i} \in {v_{parent}}\};$
		\State ${v_{parent}} \to (SI{G_v} \to sig) = s;$
		\State UnionSignature(${v_{parent}}$);
	\EndIf
\EndProcedure
\end{algorithmic}

The $Algorithm1$ creates the S-tree from the signature file $S$. Each signature $sig \in S$ is inserted into the most appropriate leaf. If the leaf is full, it would be split. Then, the S-tree grows in the direction of the root. 

The S-tree has many balanced branches. With each node of the S-tree is traversed in the best similarity measure direction with EMD distance The height of S-tree is $h = \left\lceil {{{\log }_m}n - 1} \right\rceil$. So, the cost of the query process on S-tree is $k \times M \times \left\lceil {{{\log }_m}n - 1} \right\rceil$, where $k$ is the length of each signature, $m$ is the minimum number of signatures, $M$ is the maximum number of signatures of a node in the S-tree. However, if the appropriate leaf is full, it would be split based on $\alpha  - seed$, $\beta  - seed$ operations ([4]) and the similarity measure EMD, which is done as follows:
\\\textbf{Procedure2.} Split node $v$
\\\textbf{\textit{Input:}} a node $v$
\\\textbf{\textit{Output:}} the $Stree$ (after splitting the node $v$)
\begin{algorithmic}[1]
\Procedure{SplitNode}{$v$}
	\State Create the nodes ${v_\alpha }$ and ${v_\beta }$ contain $\alpha  - seed$ and $\beta  - seed$;
	\For{$SI{G_i} \in v$}
		\If{$EMD(SI{G_i} \to sig, \beta-seed) < EMD(SI{G_i} \to sig, \alpha-seed)$}
			\State ${v_\alpha } = {v_\alpha } \cup SI{G_i};$
		\Else
			\State ${v_\beta } = {v_\beta } \cup SI{G_i};$
		\EndIf
	\EndFor
	\State ${s_\alpha } = \bigcup {sig_i^\alpha }$, with $sig_i^\alpha  \in {v_\alpha };$
	\State ${s_\beta } = \bigcup {sig_i^\beta } $, with $sig_i^\beta  \in {v_\beta };$
	\State ${v_{parent}} = {v_{parent}} \cup {s_\alpha };$
	\State ${v_{parent}} = {v_{parent}} \cup {s_\beta };$
	\If{${v_{parent}}.count > M$}
		\State SplitNode(${v_{parent}}$);
	\EndIf
\EndProcedure
\end{algorithmic}

\section{Image retrieval}
\subsection{Extracting interest regions}
In order to extract the visual features of images, we standardize the image size (i.e. converting input image in different sizes into the image with the size of $k \times k$). After that, we extract the color feature of images. Because the image based on JPEG standard is described on the YCbCr color space, we need to use YCbCr to extract the features of the images.

According to [7, 8], the Gaussian transformation by human's visual system is given by $L(x,y) = \frac{1}{{10}}{\rm{[}}6.G(x,y,{\delta _D})*Y + 2.G(x,y,{\delta _D})*Cb + 2.G(x,y,{\delta _D})*Cr]$, with $G(x,y,{\delta _D}) = \frac{1}{{\sqrt {2\pi } .{\delta _D}}}.\exp (\frac{{{x^2} + {y^2}}}{{2.\delta _D^2}})$. The intensity ${I_0}(x,y)$ for color image is calculated with equation ${I_0}(x,y) = Det(M(x,y)) - \alpha .T{r^2}(M(x,y))$, where $Det( \bullet ),Tr( \bullet )$ are Determinant and Trace of matrix, $M(x,y)$ is a second moment matrix which can be defined as $M(x,y) = \delta _D^2.G({x,y,\delta _I})*\left[ {\begin{array}{*{20}{c}}
{{L_x}^2}&{{L_x}{L_y}}\\
{{L_x}{L_y}}&{{L_y}^2}
\end{array}} \right]$, where ${\delta _I},{\delta _D}$ are the integration scale and differentiation scale, and ${L_\alpha }$ is the derivative computing the $\alpha $ direction. The interest points of color image are extracted from formula ${I_0}(x,y) > {I_0}(x',y')$ with $x',y' \in A$ and ${I_0}(x,y) \ge \theta $, where $A$ is the neighborhood of point $(x,y)$ and $\theta $ is a threshold value. Let ${O_I} = {\rm{\{ }}o_I^1,o_I^2,...,o_I^n{\rm{\} }}$ be a set of interest circles, with the centers as the interest points and the set of interest radius as ${R_I} = {\rm{\{ }}r_I^1,r_I^2,...,r_I^n{\rm{\} }}$. The values of interest radius are in ${\rm{[}}0,{{\min (M,N)} \mathord{\left/
 {\vphantom {{\min (M,N)} 2}} \right.
 \kern-\nulldelimiterspace} 2}{\rm{]}}$, where $M,N$ are the height and the width of image. They are extracted by LoG method (Laplace-of-Gaussian). For each image, the process of extracting interest points is described as follows:
\\\textbf{Algorithm2.} Extract region-of-interest in image
\\\textbf{\textit{Input:}} Image $I$, threshold $\theta $, scales ${\delta _I},{\delta _D}$
\\\textbf{\textit{Output:}} Interest region of image ${O_I} = \{ o_I^1,o_I^2,...,o_I^N\} $
\begin{algorithmic}[1]
	\State \textit{Step 1.} For each pixel $p \in I$, convert from RGB color space ($R,G,B \in {\rm{[}}0,1]$) to YCbCr color space as follows: 
	\State $\left[ {\begin{array}{*{20}{c}}
				Y\\
				{Cb}\\
				{Cr}
				\end{array}} \right] = \left[ {\begin{array}{*{20}{c}}
				{65.481}&{128.553}&{24.996}\\
				{ - 37.797}&{ - 74.203}&{112}\\
				{112}&{ - 93.786}&{ - 18.214}
				\end{array}} \right]\left[ {\begin{array}{*{20}{c}}
				R\\
				G\\
				B
				\end{array}} \right] + \left[ {\begin{array}{*{20}{c}}
				{16}\\
				{128}\\
				{128}
				\end{array}} \right]$
	\State \textit{Step 2.} Perform Gaussian transform compiled with the human visual system to calculate the $L(x,y)$ as follows:
	\\$L(x,y) = \frac{1}{{10}}{\rm{[}}6.G(x,y,{\delta _D})*Y + 2.G(x,y,{\delta _D})*Cb + 2.G(x,y,{\delta _D})*Cr]$
	\State \textit{Step 3.} Calculate the feature intensity ${I_0}(x,y)$ for color images as follows:
	\\${I_0}(x,y) = Det(M(x,y)) - \alpha .T{r^2}(M(x,y))$
	\State \textit{Step 4.} Collect the set of interest points $P$ as follows:
	\\$P = \{ p(x,y) \in I|{I_0}(x,y) > {I_0}(x',y') \wedge {I_0}(x,y) \ge \theta ,(x',y') \in A\} $
	\State \textit{Step 5.} Implement of the extraction interest regions ${O_I} = {\rm{\{ }}o_I^1,o_I^2,...,o_I^n{\rm{\} }}$ based on the set of interest points $P$.
	\\\Return ${O_I} = {\rm{\{ }}o_I^1,o_I^2,...,o_I^n{\rm{\} }}$
\end{algorithmic}
\begin{figure}
\centering
\includegraphics[height=2cm, width = 12cm]{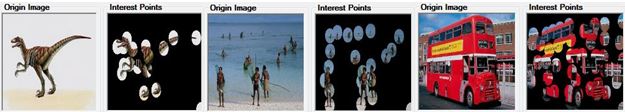}
\caption{A sample of interest regions}
\label{fig:example}
\end{figure}

\subsection{Creating binary signature of an image}
With each interest region $o_I^i \in {O_I}$ of an image $I$, the histogram is calculated on the base of the standard color range $C$. Effective clustering method relied on Euclidean measure in RGB color space classifies colors of each pixel on the image. Let $p$ be a pixel of image $I$ so we have a color vector in RGB as ${V_p} = \left( {{\rm{ }}{R_p},{\rm{ }}{G_p},{\rm{ }}{B_p}} \right)$. Let ${V_m} = \left( {{R_m},{\rm{ }}{G_m},{\rm{ }}{B_m}} \right)$ be a color vector of the set of standard color range $C$, so ${V_m} = {\rm{ }}\min \{ ||{V_p} - {V_i}||,{\rm{ }}{V_i} \in C\}$. At that time, pixel $p$ can be standardized in accordance with color vector ${V_m}$. According to experiment, the paper is used the standard color range on MPEG7 to calculate histogram for color images on COREL database. We let $o_I^i \in {O_I}$ ($i = 1,...,N$) be the interest circles of the image $I$, the histogram vector of the circle $o_I^i$ is $H(o_I^i) = \{ {H_1}(o_I^i),{H_2}(o_I^i),...,{H_{n}}(o_I^i)\} $. If we let ${h_k}(o_I^i) = \frac{{{H_k}(o_I^i)}}{{\sum\limits_j {{H_j}(o_I^i)} }}$, a standardized histogram vector will be $h(o_I^i) = \{ {h_1}(o_I^i),{h_2}(o_I^i),...,{h_{n}}(o_I^i)\} $. The binary signature describes ${h_k}(o_I^i)$ as $B_I^k = b_I^1b_I^2...b_I^{m}$ with $b_I^j = 1$ if $j = \left[ {({h_j}(o_I^i) + 0.05) \times m} \right]$, otherwise $b_I^j = 0$. So, the signature describes the interest region $o_I^i \in {O_I}$ as $Sig(o_i^I) = B_I^1B_I^2...B_I^n$. For this reason, the binary signature of the image $I$ is ${S_I} = \bigcup\nolimits_{i = 1}^N {Sig(o_I^i)} $. 
The process of creating binary signatures for color images is described as follows:
\\\textbf{Algorithm3.} Creating binary signature of image $I$
\\\textbf{\textit{Input:}} Image $I$, the look-up table of color vectors $C = ({V_1},{V_2},...,{V_n})$
\\\textit{\textbf{Output:}} Binary signature ${S_I}$ of image $I$
\begin{algorithmic}[1]
	\State \textit{Step 1.} 
	\State ${O_I}$ = Algorithm2($I,\theta ,{\delta _I},{\delta _D}$); (i.e. ${O_I} = \{ o_I^1,o_I^2,...,o_I^N\} $)
	\State Initialize binary signature ${S_I} = B_1^0B_2^0...B_n^0$, with $B_j^0 = b_1^0b_2^0...b_m^0$, $b_i^0 = 0$, $i = 1,...,m$, $j = 1,...,n$.
	\State \textit{Step 2.} 
	\State Initialize histogram vector $H(o_I^i) = \{ {H_1}(o_I^i),...,{H_n}(o_I^i)\} $
	\For {pixel $p \in o_I^i$}
		\State Calculate color vector ${V_p} = ({R_p},{G_p},{B_p})$
		\State ${V_m} = \min \{ ||{V_p} - {V_i}||,{V_i} \in C\} $
		\State ${H_m}(o_I^i) = {H_m}(o_I^i) + 1$
	\EndFor
	\State \textit{Step 3.} 
	\State Initialize vector $h(o_I^i) = \{ {h_1}(o_I^i),...,{h_n}(o_I^i)\} $
	\For {${h_k}(o_I^i) \in h(o_I^i)$}
			\State ${h_k}(o_I^i) = \frac{{{H_k}(o_I^i)}}{{\sum\limits_j {{H_j}(o_I^i)} }}$
	\EndFor
	\State \textit{Step 4.}  
	\For {${h_k}(o_I^i) \in h(o_I^i)$}
		\For {i=1..m}
			\If{$j = \left[ {({h_j}(o_I^i) + 0.05) \times m} \right]$}
				\State $b_I^j = 1$
			\Else 
				\State $b_I^j = 0$
			\EndIf
		\EndFor
		\State $B_I^k = b_I^1b_I^2...b_I^m$
	\EndFor
	\State $Sig(o_i^I) = B_I^1B_I^2...B_I^n$.
	\State \textit{Step 5.} 
	\State ${S_I} = {S_I} \vee Sig(o_I^i)$
	\State ${O_I} = {O_I}\backslash \{ o_I^i\} $
	\If {${O_I} \ne \emptyset $}
		\State To go back \textit{Step 2.}
	\Else 
	\State \textbf{return} ${S_I}$
	\EndIf
\end{algorithmic}

The \textit{Algorithm3} creates a binary signature of image $I$. At \textit{Step1} basing on \textit{Algorithm2}, it extracts the interest regions ${O_I} = \{ o_I^1,o_I^2,...,o_I^N\} $. At \textit{Step2}, it calculate the histogram vector $H(o_I^i) = \{ {H_1}(o_I^i),...,{H_n}(o_I^i)\} $ based on interest regions $o_I^i \in {O_I}$ with the set of standard color $C$.  At \textit{Step3}, it standardize vector $h(o_I^i) = \{{h_1}(o_I^i),...,{h_n}(o_I^i)\} $. At \textit{Step4}, it creates the binary signature for ${h_k}(o_I^i)$ as $B_I^k = b_I^1b_I^2...b_I^m$ with $b_I^j = 1$ if $j = \left[ {({h_j}(o_I^i) + 0.05) \times m} \right]$, otherwise $b_I^j = 0$. At that time, the signature describes the interest region $o_I^i \in {O_I}$ as $Sig(o_i^I) = B_I^1B_I^2...B_I^n$. At \textit{Step 5}, it creates the binary signature of image $I$ as ${S_I} = \bigcup\nolimits_{i = 1}^N {Sig(o_I^i)} $.

\subsection{Image retrieval algorithm}
After storing the signatures and identifications of the corresponding images on the S-tree, the query process finds out the binary signatures of similarity images on the base of traversing S-tree. After finding the image's signatures and relying on identifications of images, we can find out a set of similarity images corresponding to a query image. For this reason, the problem needs to find out the signatures of the images and the corresponding identifications. This query process is performed by proposed algorithm as follows:
\\\textbf{Algorithm4.} Query image in S-tree
\\\textbf{\textit{Input:}} query signature $sig$ and $S-tree$
\\\textbf{\textit{Output:}} Set of image signatures and identifications
\begin{algorithmic}[1]
		\State $v = root$; $SIGOUT = \emptyset $; $Stack = \emptyset $;
		\State Push(Stack, v);
		\While { \textbf{not} Empty($Stack$)} 
			\State $v$ = Pop($Stack$);
			\If{$v$ \textbf{is not} $Leaf$}
				\For {$SIG_i \in  v$ and $SIG_i \to sig  \wedge sig = sig $}
					\State $EMD(SI{G_0} \to sig,sig)$ = $\min \{EMD(SI{G_i} \to sig,sig)|SI{G_i} \in v \}$;
					\State Push(Stack, $SIG_0 \to next$);
				\EndFor
			\Else
				\State $SIGOUT = SIGOUT \cup \{ \langle SI{G_i} \to sig,oi{d_i}\rangle |SI{G_i} \in v\}$;
			\EndIf
		\EndWhile
	\\\Return $SIGOUT$;
\end{algorithmic}

Searching process is done similarly to traversing one the S-tree. So, the cost of the query process on the S-tree is $k \times M \times \left\lceil {{{\log }_m}n - 1} \right\rceil$ where $k$ is the length of each signature, $m$ is the minimum number of signatures, $M$ is the maximum number of signatures of a node in the S-tree.

\section{Experiments}
\subsection{Image retrieval model}
The experimental process includes two phases. The first one performs pre-processing to convert image data into binary signatures and puts them into S-tree based on the similarity measure EMD. The second one performs query process corresponding to a query image which can be converted into a binary signature and queried on the S-tree on the base of similarity measure EMD. After finding the signatures of similarity images, we find out similarity images and arrange them with similarity measure EMD.
\begin{figure}
\centering
\includegraphics[height=8cm, width = 10cm]{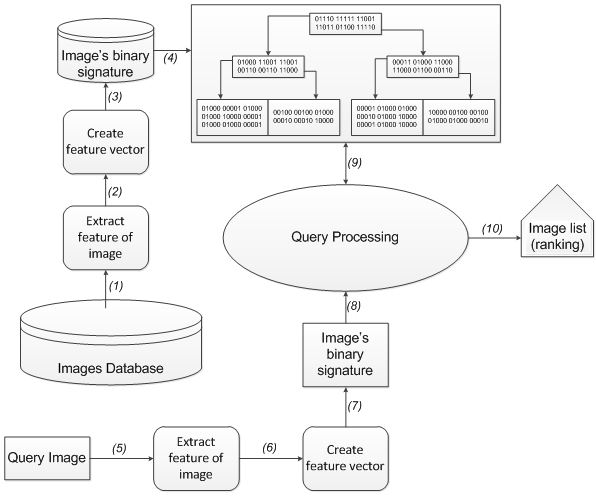}
\caption{Model of region-based image retrieval system}
\end{figure}

\textbf{\textit{Phase 1:}} \textit{Perform pre-processing}

\textit{Step 1.} Extract the features of images in the database to form feature vectors. 

\textit{Step 2.} Convert feature vectors of images into binary signatures.

\textit{Step 3.} Calculate in turn the EMD distance of image's signatures and insert them into the S-tree.

\textbf{\textit{Phase 2: }}\textit{Perform query}

\textit{Step 1.} With each query image, we extract the feature vector and convert it into binary signature.

\textit{Step 2.} To implement the binary signature retrieval process on S-tree including image's signatures, we find out similarity images at each leaf node of tree through the similarity measure EMD. 

\textit{Step 3.} After having similarity images, we arrange them in order of the similarity from high to low level and give the list of images. Then, we arrange them on the basis of similarity measure EMD.

\subsection{Experimental Results}
The experimental process is queried on COREL sample data [12] including 10,800 images which can be divided into 80 different subjects. With each query image, we retrieve images on COREL data as so as find out the most similar ones to the query image. Then, we can compare to a list of the subjects of images to evaluate the accurate method.
\begin{figure}
\centering
\includegraphics[height=5cm, width = 10cm]{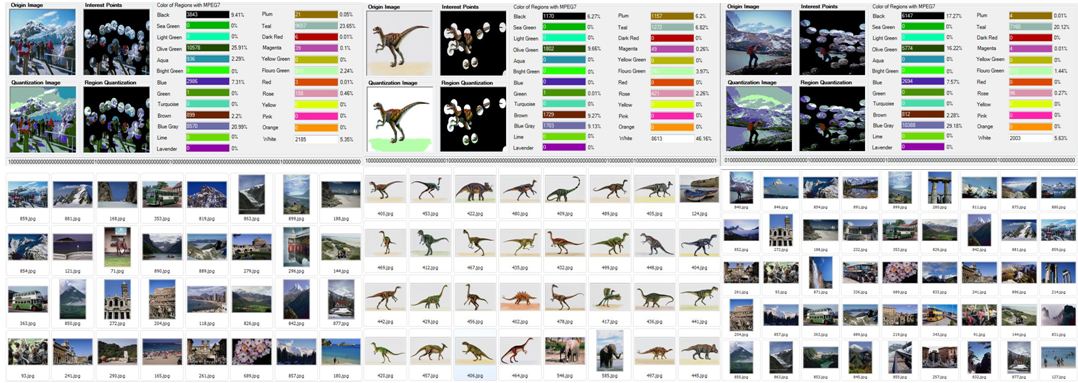}
\caption{Some results of image retrieval system}
\end{figure}
\begin{figure}
\centering
\includegraphics[height=2.5cm, width = 5.5cm]{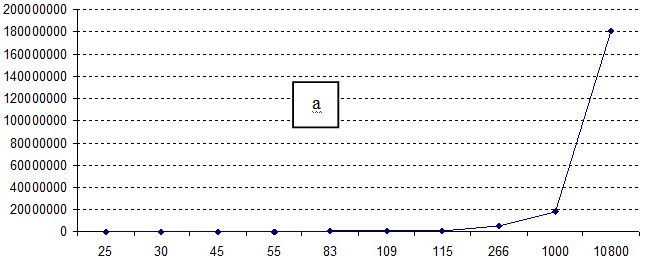}
\includegraphics[height=2.5cm, width = 5.5cm]{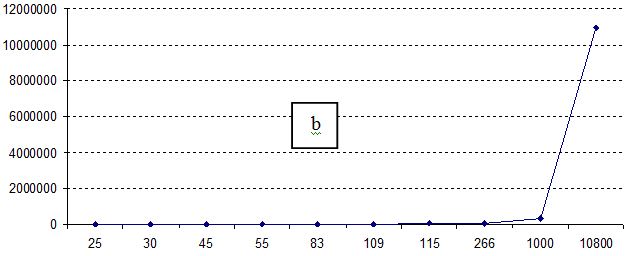}
\caption{(a)-number of operators to create S-tree; (b)-the time (milliseconds) to create S-tree}
\end{figure}
\begin{figure}
\centering
\includegraphics[height=2.5cm, width = 5.5cm]{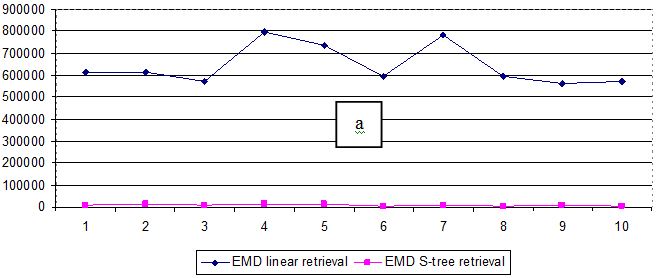}
\includegraphics[height=2.5cm, width = 5.5cm]{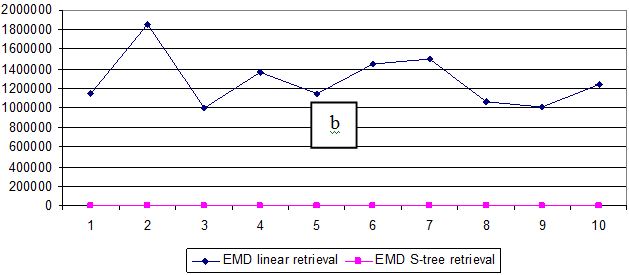}
\caption{(a)-number of operators to query image; (b)-the time (milliseconds) to query image}
\end{figure}
\begin{figure}
\centering
\includegraphics[height=2.5cm, width = 5.5cm]{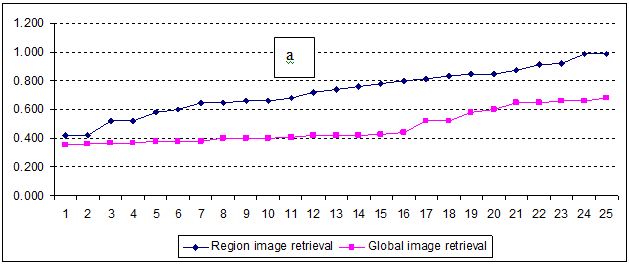}
\includegraphics[height=2.5cm, width = 5.5cm]{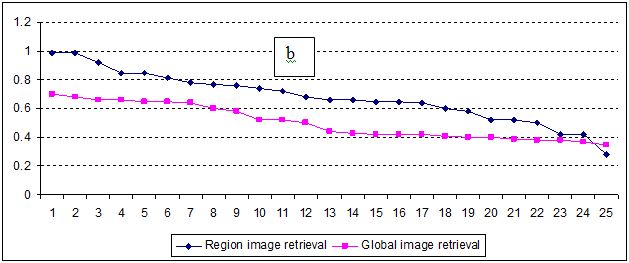}
\caption{(a)-Recall; (b)-Precision}
\end{figure}

\section{Conclusion}
The paper gives the evaluation method of similarity between two images on the base of binary signatures. At the same time, it simulates the experimental application on classification of image data of COREL. The experiment shows the evaluation method based on binary signatures and S-tree which can speed up querying similarity images as quickly as possible compared with the linear query. However, the use of feature color gives an inaccurate results with the meaning of image's contents. Therefore, the next development extracts the feature objects on images. So, it builds a binary signature to describe objects as well as the content of images. At the same time, it creates a data structure to describe the relationship on the base of similarity measure between the images.

\vspace{2cm}

\noindent\textbf{Thanh The Van}\\
Faculty of Information Technology\\
Hue University of Sciences, Hue University\\
77 Nguyen Hue street\\
Hue city\\
Vietnam\\
{\tt vanthethanh@gmail.com}\\

\noindent\textbf{Thanh Manh Le}\\
Hue University\\
03 Le Loi street\\
Hue city\\
Vietnam\\
{\tt lmthanh@hueuni.edu.vn}

\end{document}